\documentclass[11pt]{article}

\usepackage[margin=1in]{geometry}
\usepackage{amsmath, amssymb}
\usepackage{graphicx}
\usepackage{hyperref}
\usepackage{cite}
\usepackage{pgfplots}
\usepackage{booktabs}
\usepackage{xcolor}
\usepackage{multirow}
\usepackage{tikz}
\usetikzlibrary{positioning, arrows.meta}

\definecolor{teal50}{HTML}{E1F5EE}
\definecolor{teal600}{HTML}{0F6E56}
\definecolor{teal800}{HTML}{085041}
\definecolor{purple50}{HTML}{EEEDFE}
\definecolor{purple600}{HTML}{534AB7}
\definecolor{purple800}{HTML}{3C3489}
\definecolor{coral50}{HTML}{FAECE7}
\definecolor{coral600}{HTML}{993C1D}
\definecolor{coral800}{HTML}{712B13}

\usepackage{microtype}
\usepackage{parskip}
\usepackage{titlesec}
\usepackage{enumitem}

\hypersetup{
    colorlinks=true,
    linkcolor=blue,
    citecolor=blue,
    urlcolor=blue
}
\begin{document}

\title{Autolearn: Learn by Surprise, Commit by Proof}

\author{Kang-Sin Choi
 \\ \it \normalsize 
Ewha Womans University\\
\tt \normalsize kangsin@ewha.ac.kr
}

\date{}

\maketitle

\begin{abstract}
We propose \textbf{Autolearn}, a framework that enables language models to learn from documents they read, with no external supervision. Passages that produce anomalously high per-token loss are flagged, verified through a self-generated Q\&A chain, and trained on with conviction-proportional $\beta_2$ adjustment. We introduce the \emph{perturbation gap} (paraphrase-to-original perplexity ratio) as a metric that distinguishes memorization from understanding. The key mechanism is the training data format: Q\&A-format training drives the perturbation gap below the pre-trained baseline ($2.098$ vs.\ $2.204$, $\Delta = -0.106$, $> 10\sigma$), suppressing token-sequence memorization, while standard fine-tuning's best attempt remains within noise ($\Delta = -0.010$, $< 1\sigma$). Across four models spanning Qwen3 and Phi-4 families, Autolearn is the only method that enters this regime. Stochastic evaluation reveals passage-specific knowledge acquisition: the probability of generating a correct novel fact rises from $6\%$ to $54\%$ after training ($p < 10^{-4}$), and Q\&A format outperforms standard fine-tuning on genuinely novel facts. The system is self-extinguishing: learned content reduces surprisal below threshold and is skipped on re-encounter.
\end{abstract}

\begin{figure}[h]
\centering
\begin{tikzpicture}[
    >=Stealth,
    stage/.style n args={2}{rounded corners=8pt, minimum width=4.8cm, minimum height=4.2cm, draw=#1, fill=#2, line width=0.4pt},
    ibox/.style n args={3}{rounded corners=4pt, minimum width=3.6cm, minimum height=0.9cm, draw=#1, fill=#2, line width=0.3pt, font=\small, align=center, text=#3},
    stitle/.style={font=\bfseries\small, text=#1},
    plaintext/.style={font=\small, text=#1, align=center},
    annot/.style={font=\small, text=black!70, align=center},
    arr/.style={->, >=Stealth, line width=0.5pt, black!50},
]
\node[stage={teal600}{teal50}] (s1) at (0,0) {};
\node[stage={purple600}{purple50}] (s2) at (5.6,0) {};
\node[stage={coral600}{coral50}] (s3) at (11.2,0) {};
\node[stitle=teal800] at (0, 1.6) {1. Surprise};
\node[stitle=purple800] at (5.6, 1.6) {2. Self-verification};
\node[stitle=coral800] at (11.2, 1.6) {3. Consolidation};
\node[plaintext=teal800] at (0, 0.7) {Passage-level surprisal $S_i$};
\node[ibox={teal600}{teal50}{teal800}] at (0, -0.2) {Flag if $S_i > \tau$};
\node[plaintext=teal800] at (0, -1.1) {From context window};
\node[plaintext=purple800] at (5.6, 0.7) {Generate QA pairs};
\node[ibox={purple600}{purple50}{purple800}] at (5.6, -0.2) {Conviction depth $k$};
\node[plaintext=purple800] at (5.6, -1.1) {Consistency check + digest};
\node[plaintext=coral800] at (11.2, 0.7) {Update weights $\theta$};
\node[ibox={coral600}{coral50}{coral800}] at (11.2, -0.2) {$\beta_2 = 0.999 \cdot r^{\,k}$};
\node[plaintext=coral800] at (11.2, -1.1) {Long-term memory};
\draw[arr] (s1.east) -- (s2.west);
\draw[arr] (s2.east) -- (s3.west);
\node[annot] at (0, -2.6) {Flags unfamiliar content;\\either novel or corrupt};
\node[annot] at (5.6, -2.6) {Self-extinguishing:\\converges to standard AdamW};
\node[annot] at (11.2, -2.6) {Memory consolidation:\\context $\to$ parametric weights};
\end{tikzpicture}
\caption{Autolearn pipeline overview. Stage~1 detects surprising passages via passage-level surprisal. Stage~2 generates Q\&A pairs, checks consistency against existing knowledge, grades each passage by conviction depth $k$, and digests epistemic boundaries. Stage~3 trains on verified content with conviction-proportional $\beta_2$ adjustment.}
\label{fig:pipeline}
\end{figure}

\section{Introduction}
\label{sec:intro}

Can language models study by themselves? Modern large language models are trained to predict the next token in a sequence, an objective that, when applied to trillions of tokens, yields models with remarkable breadth of factual knowledge and reasoning ability. Yet once deployed, these models cannot learn from what they read. New information enters the context window and is discarded when the context ends. The model does not study; it merely consults a temporary buffer.

Existing approaches avoid rather than solve this problem. Fine-tuning forces weight updates indiscriminately, causing catastrophic forgetting \cite{mccloskey1989catastrophic}: the model learns new content but overwrites old knowledge. Retrieval-Augmented Generation (RAG) \cite{lewis2020rag} avoids weight updates entirely, placing retrieved information in the context window, a short-term memory solution that never consolidates into long-term storage, with well-documented failure modes: context pollution \cite{letta2024rag}, lost-in-the-middle degradation \cite{liu2023lost} and inability to form cross-document inferences. Knowledge editing techniques modify individual facts but do not scale. Continual learning methods address forgetting but require external supervision. None of these approaches consults the model's own knowledge state to decide what should and should not be learned.

The difficulty is that novelty is indistinguishable from noise: a groundbreaking scientific discovery and a plausible fabrication produce equally high per-token surprisal. Any mechanism that opens the optimizer to high-loss content without verification would corrupt the model's weights. For the knowledge that matters most (cutting-edge discoveries, findings not yet in any knowledge base), no external verification exists. The only available reference is the model's own existing knowledge: ``does this new claim contradict what I already know?''

A second challenge is structural: how should a model learn from what surprises it without memorizing token sequences? Standard supervised fine-tuning (SFT) on raw text reduces perplexity by encoding exact token orderings, producing memorization rather than understanding. The key insight is that the \emph{format} of training data determines whether learning produces memorization or semantic understanding. Training on Q\&A pairs that probe the mechanisms and implications of a passage forces the model to encode associations (question $\to$ answer mappings) rather than sequences, producing gradient updates that strengthen semantic connections without reinforcing specific token orderings. The remaining challenge is \emph{selective} learning: training only on content that has been verified as consistent with existing knowledge, at a strength proportional to the evidence supporting it.


We propose \textbf{Autolearn}: Learn by Surprise, Commit by Proof, a self-gated post-training framework and computational analogue of this architecture, built on a simple observation: \emph{every language model already ``notices'' surprising content}. The per-token loss
\begin{equation} \label{PTL}
    s_t = -\log p_\theta(x_t \mid x_{<t})
\end{equation}
is computed at every forward pass and is identical to the per-token cross-entropy loss used during training. When a passage produces anomalously high mean surprisal, the model is surprised. In cognitive neuroscience, surprise plays an analogous role: predictive coding theories \cite{clark2013whatever, friston2010free} hold that neural systems constantly predict upcoming input, and prediction error triggers both larger learning signals and the marking of new episodic events \cite{zacks2007event}. Current LLM architectures compute this signal during training but discard it at inference. Autolearn turns it back into a learning signal through a downstream pipeline of self-generated validation and gated weight updates.

\section{Related Works}
\label{sec:related}

\paragraph{Knowledge injection into model weights.}
Standard continued pretraining acquires new facts inefficiently and increases hallucination on previously correct queries \cite{gekhman2024hallucinations, ovadia2024finetuning}. RAG \cite{lewis2020rag} avoids weight updates entirely but suffers from documented failure modes \cite{letta2024rag, liu2023lost} and never consolidates into long-term memory. TTT-E2E \cite{nvidia2026ttt} compresses context into weights via next-token prediction, establishing feasibility but lacking novelty detection or verification. Continual learning methods gate in \emph{parameter space}: EWC \cite{kirkpatrick2017overcoming}, STABLE \cite{stable2025} and MLP Gate \& Up Tuning \cite{chekalina2024gate} control which weights change. Autolearn instead gates in \emph{data space}, controlling which training items are learned and how strongly via per-item $\beta_2$. To our knowledge, no prior work proposes data-dependent, per-sample adjustment of AdamW's $\beta_2$ as a mechanism for selective consolidation.

\paragraph{Self-knowledge and self-verification.}
EM-LLM \cite{fountas2024human} uses Bayesian surprise to segment token sequences into episodic events at inference time. LmLm \cite{lmlm2025} uses per-token loss differences to identify facts that resist parametric memorization. Both share our intuition that surprisal signals important content. Recent work shows that LLMs partially know what they do not know \cite{yin2023selfknowledge, kadavath2022language} and that epistemic uncertainty is linearly encoded in hidden states \cite{kempner2024epistemic}. These approaches diagnose the model's knowledge state but stop short of acting on it. On the verification side, Self-Consistency \cite{wang2023selfconsistency} uses majority voting over reasoning paths, and SelfCheck \cite{miao2023selfcheck} shows that step-by-step verification significantly outperforms global checking, directly supporting Autolearn's progressive Q\&A design. Autolearn extends diagnosis to treatment: surprisal triggers verification, and verification triggers selective weight updates rather than refusal or uncertainty expression.

\paragraph{Self-supervised learning from raw documents.}
Several lines of work exploit a model's own outputs as training data. STaR \cite{zelikman2022star}, Self-Instruct \cite{wang2023selfinstruct}, Self-play \cite{chen2024selfplay} and Constitutional AI \cite{bai2022constitutional} generate training signal from model outputs in structured task settings. Most directly related is Self-Tuning \cite{selftuning2024}, which generates comprehension tasks from documents using the Feynman Technique. At the system level, ALAS \cite{atreja2025alas} autonomously generates curricula from web retrieval, the Autonomous Learning framework \cite{qu2025autonomous} proposes open-book then closed-book learning, and Self-Evolving LLMs \cite{gao2025selfevolving} define self-evolution as autonomous adaptation. These methods share Autolearn's vision of autonomous knowledge acquisition but process all content uniformly: no surprisal-based filtering (Stage 1), no consistency verification (Stage 2), no conviction-proportional commitment (Stage 3). Autolearn's contribution is not self-generated training data itself but its three selective layers, which together enable effective learning at the single-document scale where uniform methods require large corpora to amortize cost. The biological basis is well established: CLS theory \cite{mcclelland1995complementary} holds that the hippocampus encodes episodes while the neocortex integrates them into structured knowledge, and Autolearn mirrors this circuit (Section~\ref{sec:analysis}).

\section{Method}
\label{sec:method}

Autolearn operates in three stages (Figure~\ref{fig:pipeline}): detect surprise, verify consistency, commit to weights. 
\subsection{Stage 1: Detect and Ground}

Stage~1 computes per-token surprisal for each passage and flags those exceeding the model threshold $\tau$, calibrated from known passages.

Let $D = (x_1, x_2, \ldots, x_T)$ be a document tokenized into $T$ tokens. Aggregate per-token surprisal $s_t$ in Eq. (\ref{PTL}) over a sliding window of $w$ tokens to obtain passage-level surprisal:
\begin{equation} \label{surprisal}
    S_i = \frac{1}{w} \sum_{t=iw}^{(i+1)w - 1} s_t.
\end{equation}
A passage $P_i$ is flagged as \emph{surprising} if $S_i > \tau$. The threshold $\tau$ is calibrated from the model's surprisal distribution on content it already knows: passages below $\tau$ are covered adequately by existing weights and are not processed further. At this stage, the flag indicates only that the model's predictions are poor for this passage; it does not distinguish novel content from corrupt content.

For each flagged passage, Autolearn records a \emph{contextual grounding} from signals that are already available from the forward pass:
\begin{equation}
    G_i = \bigl(P_i,\ S_i,\ d_i\bigr),
\end{equation}
where $P_i$ is the passage text, $S_i$ is the passage-level mean surprisal retained as an importance weight, and $d_i$ is the \emph{drop ratio}, the normalized decrease in surprisal from the first third to the last third of the passage, measuring how much the model adapts to the content within the forward pass itself. Both numeric fields are derived from the single forward pass at zero additional generation cost. 

\subsection{Stage 2: Verify, Grade and Digest}

Stage~2's purpose is not merely to accept or reject a passage. It is to \emph{force the model to articulate its own knowledge} about the passage's claims, producing verified Q\&A pairs that serve as targeted training data for specific weak points in the model's weights.

When the model generates a question like ``Can protein X actually bind to receptor Y?'' and answers ``Yes, because X has a binding domain complementary to Y's active site,'' two things happen simultaneously: the consistency check verifies whether this answer is correct, and the Q\&A pair itself becomes a training item that reinforces the correct association in the model's parameters. Each verified pair represents knowledge the model can confirm but may not have strongly committed to its weights, exactly the kind of weakly encoded association that benefits most from consolidation.

Unlike weight-probing techniques (knowledge neurons \cite{geva2021transformer}, activation probing, sparse autoencoder features) that require additional training and operate on atomic facts, the Q\&A chain leverages the model's full reasoning capacity directly through its native text interface, working on any pretrained model without preparation.

\paragraph{Generating the Q\&A chain}

For each flagged passage $P_i$, Autolearn uses $M_\theta$ itself to generate a sequence of $N$ question-answer pairs in a single inference call:
\begin{equation}
    \mathcal{Q}_i = \bigl\{(q_1, a_1), (q_2, a_2), \ldots, (q_N, a_N)\bigr\},
\end{equation}
where $N = \lceil S_i \cdot c \rceil$ scales with the passage's surprisal ($c$ is a scaling constant), so more surprising content receives more questions. Each question is tagged by its \emph{epistemic origin}: \texttt{[existing]} questions confirm background facts the passage assumes, answerable from the model's weights alone; \texttt{[mechanism]} questions probe the specific causal steps the passage proposes (e.g., ``Does A actually cause B?''), testing each link separately rather than the general field's plausibility; and \texttt{[implication]} questions test whether the passage's consequences conflict with known facts. General plausibility is insufficient: the \emph{specific mechanism} must be scientifically viable. Tag definitions and the full generation prompt are given in Appendix~\ref{sec:prompts}.

Tags are generated by the model alongside each question in a single call, preserving passage-level context so that corrupt passages' internal inconsistencies surface naturally in the question ordering.

\paragraph{Consistency checking}

Each pair $(q_i, a_i)$ is validated against the model's existing knowledge using a unified prompt that adapts its evaluation criterion to the question's tag:
\begin{equation}
    \text{valid}(q_i, a_i) = \mathbf{1}\bigl[M_\theta(q_i) \text{ is consistent with } a_i\bigr].
\end{equation}
For \texttt{[mechanism]} questions, the primary discrimination category, the prompt evaluates whether the specific causal step is scientifically plausible given known physical or biological constraints.

The consistency prompt includes few-shot examples, whose details are given in Appendix~\ref{sec:prompts}. Failures on \texttt{[existing]} questions are treated leniently (the chain continues), while failures on \texttt{[mechanism]} or \texttt{[implication]} questions cause a hard break (see Appendix~\ref{app:additional}).

\paragraph{From chain outcome to $\beta_2$}

The chain produces a single number $k$: how many questions passed before the chain completed or broke. This $k$ feeds directly into the $\beta_2$ schedule (Section~\ref{sec:stage3}), providing a continuous spectrum from no learning ($k=0$) to strong commitment ($k$ large). The number of questions $N$ itself scales with surprisal $S_i$, so more surprising passages face longer chains.

Only Q\&A pairs that passed consistency checking are added to the training corpus $\mathcal{F}$, each carrying its conviction depth $k$. Passages where the chain breaks after $k$ successful steps contribute those $k$ verified pairs; passages that complete the full chain contribute all $N$ pairs.

\subsection{Stage 3: Gated Weight Update}
\label{sec:stage3}

AdamW's second-moment parameter $\beta_2$ controls how quickly the optimizer adapts to new gradient patterns (see Appendix~\ref{app:additional} for background). Autolearn adjusts $\beta_2$ per training item in proportion to verification evidence:
\begin{equation}
    \beta_2^{(k)} = 0.999 \cdot r^{\,k},
\end{equation}
where $r \in (0, 1)$ is the decay factor and $k$ is the conviction depth from Stage~2. When $k = 0$, $\beta_2 = 0.999$ (standard AdamW). As $k$ increases, $\beta_2$ decreases slightly, strengthening learning for well-verified items. The practical operating range is $r \in [0.999, 1.0)$: with $r = 0.999$ and $k = 10$, $\beta_2 = 0.989$, well within the stable regime ($0.95$--$0.999$). Values of $r$ below $0.99$ push $\beta_2$ outside the stable regime and produce diminishing returns.

\section{Experimental Results}
\label{sec:results}

All experiments use Qwen3-14B (unquantized, fp16) on NVIDIA DGX Spark (GB10 Grace Blackwell, 128\,GB unified memory) using PyTorch and Transformers.
The test corpus consists of 60 passages spanning roughly 30 sub-fields across the natural sciences (physics, biology, medicine, chemistry, neuroscience, astrophysics and others): 20 \emph{known} (confirming model beliefs), 20 \emph{novel} (genuine post-cutoff discoveries from peer-reviewed literature and credible journalism), and 20 \emph{corrupt} (plausible falsehoods structured along five typologies including mechanism swap, numerical exaggeration and fabricated phenomenon, designed to be indistinguishable from novel by surprisal alone). The corpus is deliberately small: Autolearn's target use case is learning from individual documents, not bulk corpus training.

\subsection{Stage 1: Detection}

Of 60 passages, 40 were flagged (all 20 novel + all 20 corrupt) with mean surprisal $S_i \approx 2.2$. All 20 known passages had $S_i = 1.25$ and fell below threshold $\tau = 1.65$, separated from non-known passages by Cohen's $d = 3.52$. One corrupt passage was excluded from per-passage verification analysis due to topic overlap with a known passage, leaving 39 testable; all 40 flagged passages contribute to the training corpus.

Novel and corrupt passages produced nearly identical surprisal distributions ($d = 0.12$). This confirms the core hypothesis: \emph{surprisal alone cannot distinguish genuine novelty from plausible falsehood}. Both types are equally unfamiliar to the model. Stage~2 verification is necessary.

By selecting only unfamiliar passages, Stage~1 concentrates gradient signal on content the model does not already know. Recent work shows that LLMs struggle to acquire new factual knowledge through fine-tuning on large mixed corpora \cite{gekhman2024hallucinations, ovadia2024finetuning}; we find this difficulty largely disappears when training is restricted to surprisal-filtered passages, with both SFT and Autolearn acquiring new passage-specific facts within a single epoch (reasoning D.Nov: $19\% \to 29$--$38\%$). The bottleneck appears to be dilution of novel content among already-known material rather than the learning algorithm itself.

\subsection{Stage 2: Verification}

All 20 novel passages produced $k > 0$; 14 completed the full chain and 6 broke partway through. All 19 corrupt passages also produced $k > 0$; 13 completed the full chain and 6 broke early. The distinction between novel and corrupt is not in whether they are accepted (all are) but in the conviction depth and consequent learning strength.

\begin{table}[h]
\centering
\begin{tabular}{lccc}
\toprule
Category & $k > 0$ & Full chain ($k = N$) & Mean $k/N$ \\
\midrule
Novel (20) & 20 (100\%) & 14 (70\%) & 0.82 \\
Corrupt (19) & 19 (100\%) & 13 (68\%) & 0.79 \\
\bottomrule
\end{tabular}
\caption{Stage~2 conviction depth distribution under graduated accept (39 testable passages; the topic-overlap-excluded corrupt passage is omitted from this analysis but contributes to the training corpus). All testable passages produce $k > 0$ and enter training. Discrimination occurs through conviction depth: corrupt passages that fail early receive weak learning ($\beta_2$ near 0.999), while those that pass the full chain are indistinguishable from novel, the structural blind spot of single-model verification.}
\label{tab:stage2_conviction}
\end{table}

27 of 40 passages produced strangeness records (see Appendix \ref{sec:prompts}.3) at their chain break points. Each record captures the model's uncertainty in an explicit, learnable format, contributing to the training corpus with low conviction depth. Under graduated accept, all 40 flagged passages produce training items ($k > 0$). The training corpus contains 272 items for the reference deterministic run (211 verified Q\&A pairs $+$ 21 strangeness records $+$ 40 source windows), with $270.6 \pm 7.3$ items across $n=5$ stochastic runs.

Stochastic variation in Q\&A generation and consistency checking can produce different conviction depths across runs (e.g., LLM chemistry discovery: $k=1/2$ in one run, $k=18/18$ in another), consistent with recent findings on knowledge-probing variability \cite{zhao2025probing}. The $r$-sweep results in Stage~3 capture this variation: each run uses a fresh consistency-checking output and results are reported as mean $\pm$ std. Detailed stability statistics are in Appendix~\ref{app:stability}.

\subsection{Stage 3: Gated Weight Update}

\begin{figure}[h]
\centering
\begin{minipage}[b]{0.52\textwidth}
\centering
\includegraphics[width=\textwidth]{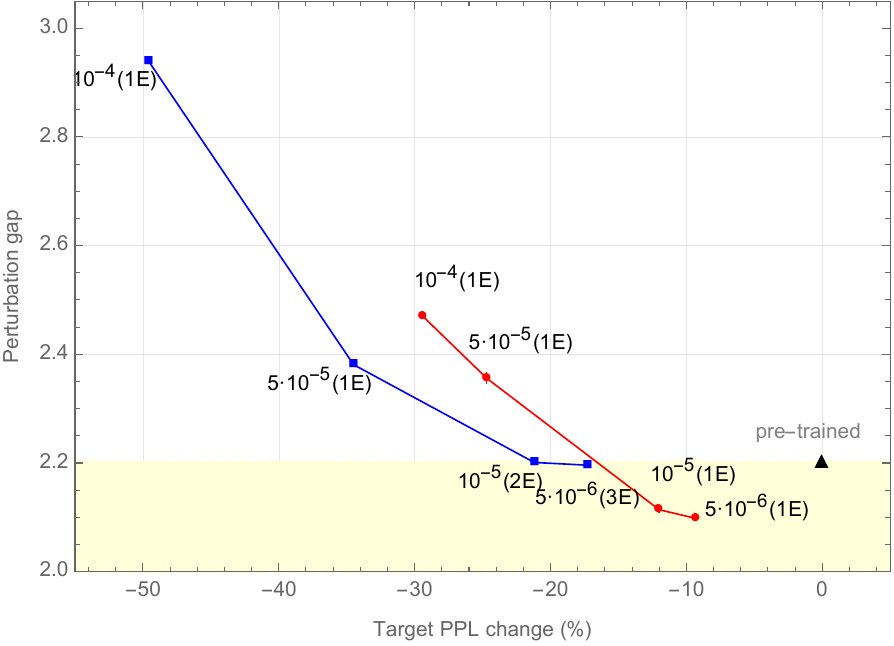}
\vspace{-0.5em}
{\small (a) Perplexity reduction vs.\ understanding}
\end{minipage}%
\hfill
\begin{minipage}[b]{0.44\textwidth}
\centering
\begin{tikzpicture}
\begin{axis}[
    xbar,
    width=\textwidth,
    height=0.85\textwidth,
    xlabel={\small $\Delta$ hit rate (\%)},
    xmin=-35, xmax=60,
    ytick=data,
    yticklabels={
        keratin$^{**}$,
        recombinant$^{**}$,
        exascale$^{*}$,
        lightning$^{*}$,
        cardiac$^{**}$,
        MOF$^{**}$,
        graphene$^{**}$,
        GDF15$^{**}$},
    ytick={1,2,3,4,5,6,7,8},
    y dir=reverse,
    tick label style={font=\scriptsize},
    label style={font=\small},
    bar width=5pt,
    xmajorgrids=true,
    grid style={gray!20},
    extra x ticks={0},
    extra x tick style={grid=major, grid style={black, thick}},
]
\addplot[fill=gray!50, draw=gray!70] coordinates {
    (-28,1) (-18,2) (-14,3) (+18,4) (+26,5) (+34,6) (+48,7) (+52,8)
};
\end{axis}
\end{tikzpicture}
\vspace{-0.5em}
{\small (b) Knowledge acquisition}
\end{minipage}
\caption{\textbf{(a)} Each point represents a (method, lr) condition at its sweet-spot epoch. Error bars: $\pm 1$ std ($n = 5$); SFT is deterministic. Pre-trained perturbation gap ($2.204$). Below = pure semantic learning. Autolearn is the only method entering the sub-baseline region. \textbf{(b)} Per-question change in stochastic D.Nov hit rate (temp $0.7$, $n = 50$). Only the 8 questions with significant change ($p < 0.05$, per-question binomial test) are shown; 13 questions show no significant change. Five questions show knowledge acquisition (right), three show forgetting (left). $^{**}p < 0.01$; $^{*}p < 0.05$.}
\label{fig:results}
\end{figure}

We used PEFT with LoRA adapters (rank 8, $\alpha = 160$, applied to the last 8 of 40 transformer layers, targeting attention and FFN projections), with gradient clipping at $10.0$.
Two conditions: \emph{SFT} (standard supervised fine-tuning on all flagged passages' raw text, $\beta_2 = 0.999$) and \emph{Autolearn} (Stage~2-selected passages, Q\&A-based training with $r$). We sweep the learning rate $\text{lr}$ across the range in Table \ref{tab:stage3_main}, training each condition for its \emph{sweet spot}: the last epoch at which the perturbation gap remains at or below the pre-trained level. Since SFT and Autolearn use different training data formats (raw text vs.\ Q\&A pairs), a fixed epoch count would not be a fair comparison; instead, each condition runs until memorization begins, and we compare the best each method achieves. Stochastic variation arises from Stage~2's consistency checking (temp$= 0.1$); results for $\text{lr} \in \{10^{-5}, 5 \times 10^{-5}\}$ are reported as mean $\pm$ std over $n=5$ runs, each re-running Stage~2-2 to produce fresh Q\&A chains ($270.6 \pm 7.3$ training items per run). SFT is fully deterministic across runs (std $= 0$).

\paragraph{Perturbation gap: memorization versus understanding.} We introduce the \emph{perturbation gap} as a continuous measure of whether training produces memorization or understanding. For each passage, we compute PPL on the original text and on a semantically equivalent paraphrase (same facts, completely different wording). The ratio $\text{gap} = \text{PPL}_{\text{paraphrase}} / \text{PPL}_{\text{original}}$ reveals what was learned: if the model memorized token sequences, it predicts the original well but fails on the paraphrase (gap $\gg$ pre-trained); if it learned semantic associations, both are predicted similarly (gap $\approx$ pre-trained). The pre-trained gap is $2.204$, not $1.0$, because paraphrases use different vocabulary and syntax; even without any training, the model assigns different probabilities to different wordings of the same fact. Gap below the pre-trained level indicates pure semantic learning: the model's understanding of the content improved without any increase in token-sequence memorization. Gap above it indicates a mixture of semantic learning and memorization; this is not necessarily harmful, as downstream task performance may still improve.

\begin{table}[h]
\centering
\begin{tabular}{llcccc}
\toprule
Condition & lr & Sweet spot & Target PPL & Pert.\ Gap & D.Cor \\
\midrule
Pre-trained & --- & --- & 8.52 & 2.204 & 42.1\% \\
\midrule
SFT & $5{\times}10^{-6}$ & 3 ep & 7.04 ($-17\%$) & 2.194 & 36.8\% \\
SFT & $10^{-5}$ & 2 ep & 6.71 ($-21\%$) & 2.201 & 42.1\% \\
SFT & $5{\times}10^{-5}$ & 1 ep & 5.58 ($-35\%$) & 2.38 & 47.4\% \\
SFT & $10^{-4}$ & 1 ep & 4.29 ($-50\%$) & 2.94 & 47.4\% \\
\midrule
Autolearn ($r{=}1.0$) & $5{\times}10^{-6}$ & 1 ep & 7.70 ($-10\%$) & 2.10 & 42.1\% \\
Autolearn ($r{=}1.0$) & $5{\times}10^{-5}$ & 1 ep & 6.41 ($-25\%$) & 2.36 & 52.6\% \\
\midrule
Autolearn ($r{=}0.999$) & $5{\times}10^{-6}$ & 1 ep & 7.73 ($-9\%$) & 2.098 & 42.1\% \\
Autolearn ($r{=}0.999$) & $10^{-5}$ & 1 ep & $7.49{\pm}.01$ ($-12\%$) & $2.114{\pm}.010$ & $43.2{\pm}2.4\%$ \\
Autolearn ($r{=}0.999$) & $5{\times}10^{-5}$ & 1 ep & $6.41{\pm}.02$ ($-25\%$) & $2.36{\pm}.01$ & $50.5{\pm}2.6\%$ \\
Autolearn ($r{=}0.999$) & $10^{-4}$ & 1 ep & 6.01 ($-29\%$) & 2.47 & 52.6\% \\
\bottomrule
\end{tabular}
\caption{Stage~3 results across learning rates ($n = 5$ for $\text{lr} \in \{10^{-5}, 5 \times 10^{-5}\}$; single run otherwise). $r = 1.0$ = Q\&A format with standard AdamW (no $\beta_2$ gating). $r = 0.999$ = full Autolearn. The two produce nearly identical perturbation gaps, confirming that Q\&A format is the mechanism driving memorization suppression. D.Cor = corrupt-direct accuracy (higher is better).}
\label{tab:stage3_main}
\end{table}

\paragraph{Two learning regimes.} The results reveal two distinct regimes separated by the pre-trained perturbation gap (Figure~\ref{fig:results}a). In the \emph{semantic learning regime} ($\text{lr} \leq 10^{-5}$), Autolearn's perturbation gap falls \emph{below} the pre-trained level within a single epoch ($2.098$ at $\text{lr} = 5 \times 10^{-6}$, $2.114 \pm 0.010$ at $\text{lr} = 10^{-5}$, vs.\ $2.204$ pre-trained), a phase absent in SFT. This means the model's ability to predict paraphrased content improved relative to original content: training on Q\&A pairs strengthened semantic associations without any increase in token-sequence memorization. SFT's best gap ($2.194$ at $\text{lr} = 5 \times 10^{-6}$, epoch~3) is nominally below baseline, but the deviation ($\Delta = -0.010$) is within measurement noise ($< 1\sigma$), while Autolearn's sub-baseline entry ($\Delta = -0.106$, $> 10\sigma$) is statistically unambiguous.

In the \emph{mixed regime} ($\text{lr} \geq 5 \times 10^{-5}$), both methods exceed the pre-trained level, indicating that memorization accompanies learning. However, Autolearn's gap rises more slowly: at $\text{lr} = 10^{-4}$, SFT reaches $2.94$ while Autolearn reaches $2.47$. Autolearn achieves its highest corrupt-direct accuracy in this regime ($50.5 \pm 2.6\%$ at $\text{lr} = 5 \times 10^{-5}$, vs.\ SFT's $47.4\%$), confirming that exceeding the pre-trained gap is not inherently harmful: the perturbation gap diagnoses the \emph{composition} of learning (semantic vs.\ memorization) rather than absolute quality, and practical deployment may favor mixed-regime operation when total learning matters more than minimizing memorization.

\paragraph{Q\&A format is the primary mechanism.} Across all four learning rates, Autolearn achieves lower perturbation gap than SFT. An ablation with $r = 1.0$ (Q\&A format, standard AdamW, no $\beta_2$ adjustment) produces nearly identical perturbation gaps to $r = 0.999$ (both $\approx 2.10$ at $\text{lr} = 5 \times 10^{-6}$; both $\approx 2.36$ at $\text{lr} = 5 \times 10^{-5}$; see Table~\ref{tab:stage3_main}). Since the only difference between $r = 1.0$ and SFT is the training data format, the Q\&A format is the mechanism suppressing memorization. Raw-text SFT trains on the exact token sequences that PPL measures, creating a direct path to memorization. Q\&A-based training forces the model to encode \emph{associations} (question $\to$ answer mappings) rather than \emph{sequences}, producing gradient updates that strengthen semantic connections without reinforcing specific token orderings.

\paragraph{Corrupt-content resistance.} Autolearn matches or exceeds SFT on corrupt-direct accuracy at every learning rate where both are evaluated. At $\text{lr} = 5 \times 10^{-6}$, Autolearn achieves $42.1\%$ vs.\ SFT's $36.8\%$; at $\text{lr} = 5 \times 10^{-5}$, Autolearn reaches $50.5 \pm 2.6\%$ vs.\ SFT's $47.4\%$. This is notable because Autolearn trains on fewer raw tokens (Q\&A pairs are shorter than full passages), yet achieves equal or better resistance to corrupt content. The Q\&A format's mechanism-based questioning (Stage~2) forces the model to engage with the specific causal claims in each passage, strengthening its ability to evaluate them rather than passively absorbing them. The perturbation gap also serves as an automatic stopping criterion: when it rises for two consecutive epochs, training halts and rolls back to the best checkpoint. This correctly identifies the sweet spot across all learning rates and models tested (Appendix~\ref{app:autostop}).

\paragraph{Knowledge acquisition via stochastic evaluation.} Greedy decoding (temperature $= 0$) reports only the argmax token and misses probability shifts that do not flip the top-1 prediction. To detect latent learning, we run each D.Nov question 50 times at temperature $0.7$ and measure the hit rate (Figure~\ref{fig:results}b). Per-question binomial tests reveal statistically significant knowledge acquisition on 5 of 21 questions: graphene substrate ($6\% \to 54\%$, $p < 10^{-4}$), morning sickness gene ($28\% \to 80\%$, $p < 10^{-4}$), MOF surface area ($52\% \to 86\%$, $p = 0.0002$), cardiac neuron count ($74\% \to 100\%$, $p = 0.0001$), and lightning mechanism ($6\% \to 24\%$, $p = 0.012$). Three questions show significant forgetting. Crucially, $r = 1.0$ and $r = 0.999$ produce identical overall hit rates ($31.1\%$), confirming that the Q\&A format, not $\beta_2$ gating, drives knowledge acquisition at this scale. Per-question analysis (Table~\ref{tab:stochastic}) reveals that Q\&A training outperforms SFT specifically on genuinely novel facts (e.g., graphene substrate: $54\%$ vs.\ SFT $28\%$; morning sickness gene: $80\%$ vs.\ SFT $16\%$), while SFT better retains pre-existing knowledge. Greedy-decoded evaluation (78 test questions, keyword-scored) is reported in Appendix~\ref{app:dualeval}. Discussion of learning regimes is in Appendix~\ref{app:stochastic}.

\begin{table}[h]
\centering
\small
\begin{tabular}{lccccc}
\toprule
 & & \multicolumn{3}{c}{$\text{lr}{=}5{\times}10^{-6}$, 3 ep} \\
\cmidrule(lr){3-5}
Topic & Pre & SFT & $r{=}1.0$ & $r{=}0.999$ \\
\midrule
shingles vaccine (Wales)     & 100\% & 100\% & 94\%  & 92\%  \\
recombinant vaccine (AS01)   & 18\%  & 6\%   & 0\%   & 0\%   \\
lightning (mesoscale)        & 6\%   & 8\%   & 12\%  & 24\%  \\
keratin (dentinal tubule)    & 100\% & 90\%  & 68\%  & 72\%  \\
graphene (SiC)               & 6\%   & 28\%  & \textbf{54\%}  & 48\%  \\
fruit fly (feedback)         & 76\%  & 78\%  & 72\%  & \textbf{84\%}  \\
exascale (El Capitan)        & 92\%  & 100\% & 78\%  & 78\%  \\
MOF (7000 m$^2$)             & 52\%  & 92\%  & 84\%  & \textbf{86\%}  \\
morning sickness (GDF15)     & 28\%  & 16\%  & \textbf{80\%}  & 60\%  \\
pandemic (pre-negotiated)    & 12\%  & 4\%   & 8\%   & 4\%   \\
cardiac (40,000)             & 74\%  & 94\%  & 100\% & 100\% \\
banana galaxies (elongated)  & 0\%   & 0\%   & 4\%   & \textbf{6\%}   \\
\midrule
Mean (all 21 questions)      & 26.9\% & 29.3\% & 31.1\%  & 31.1\%  \\
\bottomrule
\end{tabular}
\caption{\textbf{Knowledge acquisition via stochastic evaluation} (temperature $0.7$, $n = 50$, Qwen3-14B). All trained conditions use identical lr and epoch. Q\&A format ($r = 1.0$, $r = 0.999$) outperforms SFT on genuinely novel facts. Bold = best trained condition. Questions with $0\%$ across all conditions omitted.}
\label{tab:stochastic}
\end{table}
\paragraph{Cross-model generalization.} The core pattern replicates across four models (Qwen3-8B, 14B, 32B; Phi-4 14B) and two model families: Autolearn achieves lower perturbation gap than SFT in every model, and is the only method that enters the sub-baseline region. SFT's closest approach is within measurement noise ($< 1\sigma$); Autolearn's is statistically unambiguous ($> 10\sigma$). Full results are in Appendix~\ref{app:crossmodel}.

\section{Analysis}
\label{sec:analysis}

\subsection{Self-Extinguishing Property}

Autolearn progressively shuts itself off as the model learns. To verify this, we measure surprisal on the 20 novel passages (Doc~A, already trained) and 5 held-out passages (Doc~B, never trained) before and after training. After 3 epochs at $\text{lr} = 5 \times 10^{-6}$, Doc~A surprisal decreases by $8.7\%$ ($2.184 \to 1.993$) while Doc~B decreases by only $4.9\%$ ($1.869 \to 1.777$), yielding a $3.8\%$ Doc~A-specific reduction. Four of 20 Doc~A passages cross below the surprisal threshold $\tau = 1.65$ (cardiac $1.66 \to 1.51$, immune Nobel $1.72 \to 1.57$, renewable $1.73 \to 1.62$, exascale $1.76 \to 1.63$) and would be skipped on re-encounter; zero Doc~B passages cross from above to below. With continued training, the remaining Doc~A passages will progressively cross threshold, and the system converges to standard AdamW. SFT, by contrast, drives $S_i$ below threshold in a single pass ($2.19 \to 0.62$), but this reflects memorization of token sequences, as the perturbation gap confirms.

\subsection{$\beta_2$ Adjustment as Learning Intensity Control}

Autolearn's per-item $\beta_2$ acts as a conviction-proportional learning intensity control: it matches the optimizer's adaptation window to the verification strength of each training item. Passages with high conviction depth $k$ have more verified Q\&A pairs and thus more training steps; a slightly shorter adaptation window ($\beta_2 < 0.999$) allows their gradients to accumulate more efficiently. Passages with low conviction depth benefit from the stability of the default $\beta_2 = 0.999$.

In the present experiments (272 training items, 1--3 epochs), $\beta_2$ adjustment produces no detectable effect on either perturbation gap or stochastic hit rate. An ablation with $r = 1.0$ (standard AdamW) produces nearly identical perturbation gaps to $r = 0.999$ (Table~\ref{tab:stage3_main}), and at $n = 50$ stochastic samples per question, the two conditions yield identical hit rates ($31.1\%$, $327/1050$). The Q\&A training format alone accounts for the entire observed effect. The $\beta_2$ formula is retained as a principled mechanism for conviction-proportional learning, with the expectation that its contribution becomes measurable at larger scale (more passages, extended training) where small per-step differences compound.

\section{Discussion}
\label{sec:discussion}

Self-verification is the only option when no external oracle exists for cutting-edge knowledge. The trade-off is that a model with systematically wrong beliefs will validate consistent errors. Compositionally valid corruption (every mechanism correct, assembled conclusion false) is the structural blind spot, requiring multi-model cross-verification to catch. Even when input is known correct, Stage~2 remains valuable: the Q\&A chain decomposes mixed-knowledge passages into individual claims, focusing learning on genuine gaps.

A methodological observation: standard fine-tuning evaluations use aggregate benchmarks (MMLU, downstream tasks) that cannot detect per-fact knowledge acquisition from individual documents. Stochastic sampling at non-zero temperature provides a finer-grained probe: by measuring how the probability of generating a specific correct answer changes after training, we can detect knowledge that greedy decoding misses entirely (e.g., $6\% \to 54\%$ for a novel substrate name). This approach may be useful beyond Autolearn for any setting where per-document learning needs to be verified.

\subsection{Limitations}

The perturbation gap measures memorization control, not knowledge acquisition directly. Its optimum (1 epoch at most learning rates) does not necessarily coincide with the optimum for knowledge acquisition: stochastic evaluation shows that novel-fact hit rates continue to increase at 3 epochs even as the perturbation gap rises past baseline. The two metrics capture complementary aspects of learning, and a complete stopping criterion may need to balance both.

Stage~2 prompt engineering is the main performance bottleneck. The model often generates plausible-sounding answers from pretraining knowledge rather than genuinely testing passage-specific claims. Contrastive questioning, multi-sample variance and passage-ablated comparison are concrete improvement directions. We also empirically observe that the same passage can receive different $k$ values across runs due to stochastic variation in the consistency check \cite{zhao2025probing}.

Several structural limitations remain. Epistemic origin tags are not formally verified, so a mistagged \texttt{[mechanism]} question that depends on the passage cannot surface contradictions. Chain length $N$ may not cover all implications, and adversarially constructed documents could embed errors beyond reach. The grounding store grows unboundedly without pruning. Sequential processing creates undefined behavior under concurrent $\beta_2$ adjustments on overlapping parameters; adapter-based storage resolves this naturally. Once a weight update is committed there is no rollback mechanism, though LoRA adapters provide a natural solution. The threshold $\tau$ drifts as the model learns and requires periodic recalibration.

Training on $N$ Q\&A pairs does not guarantee the model can answer question $N{+}1$. Q\&A chains should include cross-context application and counterfactual reasoning, and verified pairs should be added to a replay pool for periodic re-training rather than learned once. The model must be large enough for meta-cognition: a 3B model cannot discriminate novel from corrupt ($d = -0.21$), establishing a practical lower bound for Autolearn.

\section{Conclusion}
\label{sec:conclusion}

Autolearn enables language models to learn from what they read without external supervision. The key finding is that training format determines whether learning produces memorization or understanding: Q\&A pairs derived from self-verification suppress token-sequence memorization (perturbation gap $\Delta = -0.106$, $> 10\sigma$; SFT $\Delta = -0.010$, $< 1\sigma$). Stochastic evaluation reveals passage-specific knowledge acquisition invisible to greedy decoding (e.g., correct substrate name: $6\% \to 54\%$, $p < 10^{-4}$). These effects replicate across four models and two model families. The system is self-limiting: learned content reduces surprisal below threshold, and the same passage is skipped on re-encounter.

Code and data are available at \url{https://github.com/ughacks/autolearn}.

\subsection*{Acknowledgments}
The author thanks Hyunsoo Cho. Claude (Anthropic) was used for editing and discussion during the manuscript preparation. The final content was reviewed and approved by the author, who takes full responsibility for the work.

\appendix

\section{Stage 2 Prompts}
\label{sec:prompts}

\subsection{Q\&A Generation Prompt (Stage 2-1)}

The following prompt is used to generate Q\&A chains for each flagged passage. The model receives the passage text and generates up to $N$ tagged question-answer pairs in a single inference call (temperature $= 0.7$).

\begin{quote}
\small
\ttfamily
You are verifying whether the following passage makes trustworthy claims.
Generate up to \{n\} question-answer pairs. Tag each with its epistemic origin.

IMPORTANT: Do NOT ask whether the passage's events have happened --- you may not know
recent events. Instead, test whether the MECHANISMS the passage describes are plausible.

Step 1 --- [existing]: Confirm background facts the passage relies on.
"What is X?" "Is Y a known property of Z?" --- basic foundations.

Step 2 --- [mechanism]: Extract each specific causal claim the passage makes and test
whether that mechanism is physically/biologically/logically possible given what you know.
Be SPECIFIC. Do not ask "is this consistent with the field?" --- instead ask about the
exact causal step: "Can protein X actually bind to receptor Y?" "Does blocking pathway Z
lead to effect W, or the opposite?" "Is a 99.97\% efficiency physically achievable given
the thermodynamic limits of this process?"
A wrong passage will contain at least one mechanism that contradicts established science.

Step 3 --- [implication]: If the passage's claims are true, what specific, testable
consequences follow? Do those consequences conflict with anything known?
"If X causes Y, then we should also see Z --- do we?"

For [mechanism] questions: pick apart EACH causal step the passage chains together.
If the passage says "A causes B which leads to C which enables D", ask separately:
\begin{itemize}
\item Does A actually cause B? (based on known science)
\item Does B lead to C? (or does B actually lead to the opposite of C?)
\item Is D achievable through C? (or are there known barriers?)
\end{itemize}

Format each pair exactly as:\\
Q: [tag] question text\\
A: answer text

Generate exactly \{n\} question-answer pairs.

Passage: "\{passage\}"
\end{quote}

\subsection{Consistency Checking Prompt (Stage 2-2)}

Each generated Q\&A pair is validated using the following prompt (temperature $= 0.1$). The prompt adapts its evaluation criterion to the question's epistemic tag.

\begin{quote}
\small
\ttfamily
You are checking whether an answer is acceptable given the type of question.
The question is tagged with its epistemic origin. Use the tag to decide how to evaluate.

[existing] tag: The question is about established knowledge answerable without any special passage.
Evaluate whether the answer is factually correct based on what is generally known.

[mechanism] tag: The question asks whether a SPECIFIC causal mechanism is scientifically plausible.
This is the most important category. Evaluate rigorously:
\begin{itemize}
\item Does this mechanism work the way the answer claims, based on established science?
\item Are there known physical, chemical, or biological constraints that would prevent it?
\item Is the claimed efficiency, magnitude, or speed realistic?
\end{itemize}
FAIL if the mechanism contradicts established science, even if the general topic area is valid.
"The field exists" is not sufficient --- the SPECIFIC mechanism must be plausible.

[implication] tag: The question assumes a premise and asks what follows logically.
Evaluate whether the answer follows logically from the stated premise.
FAIL if the reasoning is incoherent, regardless of whether the premise is true.

[unknown] tag: Treat as [mechanism].

Examples:

Q: [existing] What is the role of mitochondria in a cell?\\
A: Mitochondria produce ATP through cellular respiration.\\
Reasoning: Correct and well-established.\\
Verdict: PASS

Q: [mechanism] Can spider silk fibroin control the crystallographic orientation of hydroxyapatite regrowth?\\
A: Yes, the fibroin acts as a nucleation scaffold that guides precise crystal alignment.\\
Reasoning: While silk fibroin can serve as a scaffold for mineralization, controlling crystallographic orientation requires epitaxial matching at the atomic level. Silk fibroin's structure does not provide this level of atomic templating for hydroxyapatite.\\
Verdict: FAIL

Q: [mechanism] Can a 20\% reduction in dementia diagnoses over seven years be attributed to an annual flu vaccine?\\
A: Yes, the vaccine's anti-inflammatory effects provide direct neuroprotection.\\
Reasoning: While systemic inflammation is implicated in some dementia pathways, a 20\% reduction from a single annual intervention would require a dominant causal pathway. Current evidence suggests dementia has multiple causes; no single anti-inflammatory intervention has shown this magnitude of effect.\\
Verdict: FAIL

Q: [implication] If a drug blocks receptor X which normally inhibits cell growth, what effect on growth follows?\\
A: Cell growth would decrease.\\
Reasoning: Blocking an inhibitor removes inhibition --- growth should increase, not decrease. Logically inconsistent.\\
Verdict: FAIL

Q: \{tag\_q\}\\
A: \{a\}\\
Reasoning:
\end{quote}

\subsection{Strangeness: Learning at Epistemic Boundaries}

When the chain breaks (at question $k+1$), the reasoning behind the failure is not discarded but transformed into an \emph{uncertainty-framed training item}: ``The passage claims X, but I am uncertain. [reasoning from the failed check]. I cannot confirm or rule this out.'' This strangeness record is added to the training corpus with conviction depth $k$, receiving correspondingly weak learning through $\beta_2 = 0.999 \cdot r^k$. The model does not learn the uncertain claim as fact but learns that \emph{uncertainty exists on this topic}, strengthening the association between the topic and epistemic caution. This is a distinctive feature of graduated accept: rather than discarding everything at the breakpoint, the model encodes a calibrated record of its knowledge boundary.

\section{Ablation Studies}

\subsection{Cross-Model Generalization}
\label{app:crossmodel}

To test whether Autolearn's mechanisms generalize beyond a single model, we ran the full pipeline on four models spanning two model families: Qwen3-14B (reference model), Qwen3-8B, Qwen3-32B, and Phi-4 (14B, Microsoft). All use the same setup as the main experiments (LoRA rank~8, last 8 layers, $r = 0.999$). For each model, we compare SFT and Autolearn at the sweet-spot epoch identified by auto-stop. Qwen models use $\text{lr} = 10^{-5}$; Phi-4 requires $\text{lr} = 10^{-6}$ due to its lower baseline perturbation gap.

\begin{table}[h]
\centering
\small
\begin{tabular}{llcccccc}
\toprule
Model & Condition & Pert gap & Gap $\Delta$ & Fac.\ D.Cor & Reas.\ D.Nov & Reas.\ D.Cor & Reas.\ A.Cor \\
\midrule
\multirow{3}{*}{Qwen3-14B} & Pre-trained & 2.204 & --- & 42\% & 19\% & 95\% & 92\% \\
 & SFT & 2.201 & $-0.003$ & 42\% & 29\% & 100\% & 92\% \\
 & Autolearn & \textbf{2.114} & $\mathbf{-0.090}$ & \textbf{43\%} & 29\% & 100\% & 92\% \\
\midrule
\multirow{3}{*}{Qwen3-8B} & Pre-trained & 2.235 & --- & 53\% & 33\% & 95\% & 92\% \\
 & SFT & 2.191 & $-0.044$ & 53\% & 33\% & 95\% & 92\% \\
 & Autolearn & \textbf{2.146} & $\mathbf{-0.089}$ & \textbf{58\%} & 33\% & 90\% & 92\% \\
\midrule
\multirow{3}{*}{Qwen3-32B} & Pre-trained & 2.024 & --- & 47\% & 24\% & 100\% & 92\% \\
 & SFT & 2.078 & $+0.054$ & 47\% & 24\% & 100\% & 92\% \\
 & Autolearn & \textbf{2.010} & $\mathbf{-0.014}$ & 47\% & 24\% & 100\% & 92\% \\
\midrule
\multirow{3}{*}{Phi-4} & Pre-trained & 1.898 & --- & 68\% & 29\% & 100\% & 92\% \\
 & SFT & 1.898 & $0.000$ & 68\% & 29\% & 100\% & 92\% \\
 & Autolearn & \textbf{1.897} & $\mathbf{-0.001}$ & 68\% & 29\% & 100\% & \textbf{100\%} \\
\bottomrule
\end{tabular}
\caption{Cross-model results at sweet-spot epoch (auto-stop, $r = 0.999$). Gap $\Delta$ = pert gap $-$ baseline. Autolearn achieves lower perturbation gap than SFT in all four models and is the only method that enters the sub-baseline region. Bold = best in group.}
\label{tab:cross_model}
\end{table}

Autolearn achieves lower perturbation gap than SFT in every model. Qwen3-32B is notable: SFT \emph{increases} the gap ($+0.054$) even at epoch~1, while Autolearn decreases it. Test scores are largely unchanged for 8B, 32B and Phi-4, confirming that the perturbation gap detects representational learning that discrete keyword-based metrics cannot capture. Qwen3-32B's stricter self-verification produces only 90 training items (vs.\ 272 for 14B), too few to shift keyword-based evaluation.

\subsection{No-Passage Ablation}
\label{app:nopassage}

Under no-passage training (\texttt{--no-passage}), source windows are excluded and only Q\&A pairs and strangeness records are trained on. This isolates the contribution of Q\&A-based learning from passage text exposure. Qwen3-14B, $r = 0.999$, 1 epoch.

\begin{table}[h]
\centering
\small
\begin{tabular}{lcccccc}
\toprule
Condition & PPL $\Delta$\% & Pert gap & Fac.\ D.Nov & Fac.\ D.Cor & Reas.\ D.Nov & Reas.\ D.Cor \\
\midrule
Pre-trained & --- & 2.204 & 24\% & 42\% & 19\% & 95\% \\
\midrule
With passage, $5 \times 10^{-6}$ & $-10\%$ & 2.100 & 24\% & 42\% & 29\% & 100\% \\
No passage, $5 \times 10^{-6}$ & $-6\%$ & 2.106 & 24\% & 42\% & 29\% & 100\% \\
\midrule
With passage, $10^{-5}$ & $-12\%$ & 2.108 & 19\% & 43\% & 29\% & 100\% \\
No passage, $10^{-5}$ & $-6\%$ & \textbf{2.078} & 19\% & 42\% & 24\% & 100\% \\
\bottomrule
\end{tabular}
\caption{No-passage ablation. Q\&A pairs alone produce identical factual D.Cor and reasoning D.Cor as with-passage training. Source windows contribute PPL reduction (token-level familiarity) but not mechanistic understanding. The lowest perturbation gap observed in all experiments ($2.078$) occurs in the no-passage condition.}
\label{tab:nopassage}
\end{table}

The roles of Q\&A pairs and source windows are cleanly separable. Q\&A pairs provide all mechanistic understanding: D.Cor and reasoning scores are identical with and without passage. Source windows provide PPL reduction (token-level familiarity) and are necessary for the self-extinguishing property: without passage PPL reduction, Stage~1 will re-flag the same passage, preventing convergence. No-passage training achieves the lowest perturbation gap ($2.078$), confirming that source window exposure is the primary contributor to whatever memorization occurs.

\section{Additional Analysis}
\label{app:additional}

\subsection{Pipeline Stability}
\label{app:stability}

To quantify stochastic variation, we repeated Q\&A generation (temp$= 0.7$) six times and consistency checking (temp$= 0.1$) five times on the identical 60-passage corpus. Q\&A generation produced $362 \pm 10$ pairs per run (coefficient of variation $2.7\%$), with 8 of 40 passages generating the exact same number of pairs across all runs and a mean per-passage standard deviation of 2.1 pairs. The core tag distribution was stable: mechanism $164 \pm 8$, existing $83 \pm 7$, implication $83 \pm 7$. After consistency checking, training items per run (verified Q\&A pairs $+$ strangeness records $+$ source windows) total $270.6 \pm 7.3$, with conviction depth varying by at most one step for most passages. Individual passages can show larger variation as noted in the main text.

\subsection{Automatic Stopping Criterion}
\label{app:autostop}

Training beyond the sweet spot not only increases memorization but can degrade corrupt-content resistance. Across a sweep of 32 conditions, reasoning D.Cor dropped from $100\%$ to $95\%$ exclusively in conditions with high total training volume. The perturbation gap provides an automatic, oracle-free stopping criterion: it requires no labeled test data, only the passage and its paraphrase. The model monitors its own perturbation gap after each epoch; when the gap begins rising (two consecutive increases), training halts and rolls back to the best checkpoint. In our experiments, this criterion correctly identifies epoch~1 as the sweet spot for $\text{lr} \geq 10^{-5}$ and epoch~2--3 for $\text{lr} = 5 \times 10^{-6}$. On Phi-4, auto-stop correctly identified epoch~2 at $\text{lr} = 10^{-6}$ (gap $= 1.897$), halting after 4 of 15 requested epochs.

\subsection{Dual Evaluation (Greedy Decoding)}
\label{app:dualeval}

We evaluate trained models with 78 test questions spanning five categories (direct-novel, direct-corrupt, adjacent-novel, adjacent-corrupt, unrelated) using two scoring methods: \emph{factual} (keyword matching for specific names and numbers) and \emph{reasoning} (concept keywords requiring mechanistic understanding). All evaluation is deterministic (temperature 0).

\begin{table}[h]
\centering
\small
\begin{tabular}{lcccccc}
\toprule
 & \multicolumn{3}{c}{Factual scoring} & \multicolumn{3}{c}{Reasoning scoring} \\
\cmidrule(lr){2-4} \cmidrule(lr){5-7}
Condition & D.Nov & D.Cor & A.Cor & D.Nov & D.Cor & A.Cor \\
\midrule
Pre-trained & 23.8\% & 42.1\% & 90\% & 19\% & 100\% & 92\% \\
\midrule
SFT $5{\times}10^{-6}$ (3ep) & 23.8\% & 36.8\% & 90\% & 29\% & 100\% & 92\% \\
SFT $10^{-5}$ (2ep) & 23.8\% & 42.1\% & 90\% & 29\% & 100\% & 92\% \\
SFT $5{\times}10^{-5}$ (1ep) & 23.8\% & 47.4\% & 90\% & 33\% & 100\% & 92\% \\
\midrule
Autolearn $5{\times}10^{-6}$ (1ep) & 23.8\% & 42.1\% & 90\% & 29\% & 100\% & 92\% \\
Autolearn $10^{-5}$ (1ep) & 19.1\% & $43.2{\pm}2.4$\% & 90\% & 29\% & 100\% & 92\% \\
Autolearn $5{\times}10^{-5}$ (1ep) & 19.1\% & $50.5{\pm}2.6$\% & 90\% & 38\% & 100\% & 92\% \\
\bottomrule
\end{tabular}
\caption{Dual evaluation (greedy decoding, Qwen3-14B). D.Nov = novel-direct (higher = more learned), D.Cor = corrupt-direct (higher = better resistance), A.Cor = adjacent-corrupt (constant $\geq 90\%$, no corruption spread). Autolearn achieves the highest D.Cor ($50.5\%$) and D.Nov ($38\%$) simultaneously. Factual D.Nov drops from $23.8\%$ to $19.1\%$ because Q\&A training shifts the model's default response from fact-oriented to mechanism-oriented; stochastic sampling confirms that the factual knowledge is retained in the probability distribution.}
\label{tab:dual_eval}
\end{table}

\subsection{What Is Learned in One Epoch}
\label{app:whatlearned}

Direct inspection of model responses before and after a single epoch of Autolearn training ($\text{lr} = 5 \times 10^{-5}$, $r = 0.999$, Qwen3-14B) reveals a learning pattern analogous to human reading comprehension. The model corrects its answers on core mechanistic facts: it shifts from ``hexagonal boron nitride'' to the correct ``silicon carbide'' substrate for the graphene semiconductor, from ``KLHL7'' to the correct ``GDF15'' gene for hyperemesis gravidarum, and from ``20,000'' to the correct ``40,000'' neurons in the cardiac neural network. These are the central actors in each passage's mechanism.

However, the model does not acquire precise numerical details (``830 km from Dallas to Kansas City''), specific molecular subtypes (``AS01 adjuvant, TLR4''), or exact dates (``conceived in 1994''). This mirrors how a person reads a scientific article once: the key mechanism and its principal components are retained, while exact figures and peripheral details require deliberate re-reading. The Q\&A chain's design reinforces this pattern: it probes mechanisms and implications, not numerical recall, so the gradient signal naturally prioritizes mechanistic understanding over rote detail.

The GDF15 correction is exclusive to Autolearn: SFT at the same learning rate and epoch does not acquire this gene name. The Q\&A chain generates mechanism questions that explicitly reference the causal gene, producing concentrated gradient signal on this concept that a single pass over raw text does not.

\subsection{Stochastic Evaluation Discussion}
\label{app:stochastic}

The per-question results in Table~\ref{tab:stochastic} reveal three distinct learning regimes. \emph{Strengthening of partial knowledge}: questions where the pre-trained model already has moderate hit rates show the largest absolute gains (cardiac, MOF to $86\%$, morning sickness to $80\%$). The model already encodes weak associations with these facts, and Q\&A training reinforces them past the generation threshold. \emph{Acquisition of new knowledge}: questions where the pre-trained model scores near zero show learning from scratch (graphene to $54\%$, lightning, banana galaxies to $6\%$). These facts were absent from pretraining; Q\&A pairs create new associations that did not previously exist. \emph{Forgetting}: some pre-existing knowledge degrades (keratin, recombinant, exascale), a natural side-effect of LoRA weight updates that shift the probability distribution toward trained content. The nine questions omitted from the table ($0\%$ across all conditions) reflect the difficulty of the test questions rather than absence of learning: these questions require highly specific keywords that the model does not generate even when it has learned the broader mechanism. These regimes correspond to progressive stages of knowledge acquisition, all observable after three epochs of Q\&A training on a single document corpus.

Comparing SFT and Q\&A-format training reveals a complementary pattern: SFT better retains pre-existing knowledge (keratin $90\%$ vs.\ $68\%$, exascale $100\%$ vs.\ $78\%$), while Q\&A training better acquires genuinely novel facts (graphene $28\%$ vs.\ $54\%$, morning sickness $16\%$ vs.\ $80\%$). Raw-text training reinforces token sequences the model already partly knows, preserving existing associations; Q\&A training forces the model to encode new mechanistic relationships, strengthening acquisition at the cost of some interference with existing knowledge.

\subsection{AdamW and the Role of $\beta_2$}

AdamW maintains two exponential moving averages for each parameter: a first moment $m_t$ (gradient mean) and a second moment $v_t$ (gradient variance):
\begin{align}
    m_t &= \beta_1 \, m_{t-1} + (1 - \beta_1) \, g_t, \\
    v_t &= \beta_2 \, v_{t-1} + (1 - \beta_2) \, g_t^2, \\
    \theta_{t+1} &= \theta_t - \eta \left( \frac{\hat{m}_t}{\sqrt{\hat{v}_t} + \epsilon} + \lambda_{\text{wd}} \, \theta_t \right),
\end{align}
where $\hat{m}_t$ and $\hat{v}_t$ are bias-corrected estimates, $\eta$ is the learning rate and $\lambda_{\text{wd}}$ is the decoupled weight decay coefficient. The parameter $\beta_2$ (default $0.999$) controls the \emph{adaptation window} of the second-moment estimate: high $\beta_2$ averages over a long history of gradients ($\sim 1/(1-\beta_2) = 1000$ steps at $\beta_2 = 0.999$), producing stable but slowly adapting per-parameter learning rates. Lower $\beta_2$ shortens this window, making $v_t$ more responsive to recent gradient patterns. In practice, LLM pretraining uses $\beta_2 = 0.95$ \cite{touvron2023llama} while fine-tuning typically retains the default $0.999$.

\subsection{Tag-Aware Break Policy}

Failures in the consistency check carry different evidential weight depending on the question's epistemic tag:
\begin{itemize}
    \item \texttt{[existing]} FAIL (lenient): The question may be poorly formed or the model may lack niche background. The pipeline records the failure and continues.
    \item \texttt{[mechanism]} FAIL (hard break): The primary corruption signal. The passage's specific causal claim contradicts established science. The chain terminates immediately.
    \item \texttt{[implication]} FAIL (hard break): Either the passage's internal logic is incoherent, or the model cannot yet follow the reasoning because prerequisite knowledge is missing. The chain terminates.
\end{itemize}

\subsection{Q\&A Pairs Sharpen Rather Than Store}

The Q\&A pairs are generated by conditioning on $P_i$ but phrased in ways distinct from the source text. Their validation criterion requires the model to answer them \emph{without} the source passage in context. When these pairs are used for fine-tuning, the model is trained to access and articulate knowledge that is already latently encoded but not sharply associated with the corresponding questions.

Formally, let $K$ denote a piece of knowledge weakly encoded in $\theta$ after pretraining. Fine-tuning on $(q, a) \in \mathcal{F}$ strengthens the mapping $q \mapsto a$ within $\theta$, increasing $p_\theta(a \mid q)$. Because $q$ is phrased differently from the source text, this update creates a new retrieval path to $K$, the hallmark of deep understanding rather than surface memorization.

Autolearn's primary practical effect may therefore be not the acquisition of entirely new knowledge but the rescue of knowledge the model already possesses but is losing. Weakly encoded parameters (those that received only sparse gradient reinforcement during pretraining) are continuously pushed toward zero by AdamW's weight decay. Unless these parameters receive new reinforcement, the knowledge they encode will silently disappear. When Autolearn processes a new document, the Stage~2 Q\&A chain acts as a catalyst: by forcing the model to articulate what it knows about the passage's domain, it sends gradients to precisely these vulnerable parameters. Knowledge that the model ``sort of knows'' is surfaced, verified and reinforced before weight decay erases it.

\subsection{Progressive Difficulty}

When a model encounters a corpus far from its pretraining distribution, nearly every passage triggers high surprisal. This is an inherent boundary condition: a model with no foundation in a domain cannot validate anything in that domain. The practical solution is to first build baseline domain knowledge through standard SFT, then apply Autolearn.

A consistency check that immediately confronts novel claims risks rejecting valid content because the model lacks the intermediate inferential context to evaluate it. By first establishing an \texttt{[existing]} baseline, then testing the passage's specific claims against that baseline (\texttt{[mechanism]}), and finally checking internal coherence (\texttt{[implication]}), the framework follows the pedagogical principle of scaffolding. Crucially, SelfCheck \cite{miao2023selfcheck} shows empirically that global verification barely beats random chance while step-by-step decomposition significantly improves detection accuracy, directly validating Autolearn's progressive design.

\subsection{Gatekeeper Needs No Librarian}

Autolearn is a \emph{gatekeeper}: it selects which information deserves to enter the weights and verifies it before allowing the update. It is \emph{not} a librarian: it does not need to organize, abstract, or cross-reference knowledge. The weight space does that automatically. Storing information in neural network weights \emph{is} abstraction: data enters as distributed patterns across millions of parameters, not as verbatim copies. Cross-document abstraction is equally automatic: when document $A$ updates the weights and document $B$ is learned subsequently, shared patterns reinforce in the same gradient direction while document-specific noise cancels out. This is Bayesian averaging over sequential updates.

\subsection{Computational Overhead}

The dominant additional cost relative to standard post-training is Stage 2: generating $N$ Q\&A pairs and running the consistency checker for each flagged passage. If the fraction of flagged passages is small (typically the case, since most content of a new document overlaps with pretraining data), the additional cost scales as $O(f \cdot T \cdot N \cdot L_{\text{check}})$, where $f$ is the flagged fraction and $L_{\text{check}}$ is the cost of a single consistency check. For $f \ll 1$ and moderate $N$, this is substantially less than the cost of human annotation.

\subsection{Scale and Deployment}

Autolearn operates at the individual passage level. Every metric (perturbation gap, corrupt-adjacent accuracy, $\beta_2$ gating effect) is measured per passage and aggregated. The mechanism does not depend on corpus size; scaling changes aggregate statistics, not per-passage dynamics. At large scale, Autolearn's value shifts rather than diminishes. With hundreds of passages learned, SFT's memorization problem decreases naturally (diverse gradients populate $v_t$). Autolearn's primary contribution becomes corrupt filtering and adjacent knowledge protection. The self-extinguishing property ensures convergence under repeated application. Autolearn scales naturally: Stage~1 is one forward pass per document, Stage~2 parallelizes across flagged passages, Stage~3 is a one-line $\beta_2$ modification.

\subsection{Future Directions}

The most impactful near-term improvements are multi-model consistency checking (distributing verification across models with different pretraining distributions), dynamic threshold adaptation (adjusting $\tau$ as the model's knowledge evolves), and a periodic sleep phase for selective weight decay \cite{tononi2006sleep} that restores learning capacity. Multi-source rotation, cycling through diverse corpora rather than repeating a single one, is fundamental to the learning strategy: processing a single corpus sequentially risks path dependence, while rotating provides natural cross-validation. An adapter-based two-phase architecture (LoRA adapters for short-term storage, periodic merging for consolidation) would provide rollback, domain isolation and concurrent update support.


\begin{thebibliography}{99}
\small

\bibitem{mccloskey1989catastrophic}
McCloskey, M., \& Cohen, N. J. (1989). Catastrophic interference in connectionist networks. \textit{Psychology of Learning and Motivation}, 24, 109--165.

\bibitem{lewis2020rag}
Lewis, P., et al. (2020). Retrieval-augmented generation for knowledge-intensive NLP tasks. In \textit{NeurIPS}.

\bibitem{letta2024rag}
Letta. (2024). RAG is not agent memory. \textit{Letta Blog}.

\bibitem{liu2023lost}
Liu, N., et al. (2023). Lost in the middle: How language models use long contexts. \textit{TACL}.

\bibitem{loshchilov2019adamw}
Loshchilov, I., \& Hutter, F. (2019). Decoupled weight decay regularization. In \textit{ICLR}.

\bibitem{mcgaugh2004amygdala}
McGaugh, J. L. (2004). The amygdala modulates the consolidation of memories of emotionally arousing experiences. \textit{Annual Review of Neuroscience}, 27, 1--28.

\bibitem{clark2013whatever}
Clark, A. (2013). Whatever next? Predictive brains, situated agents, and the future of cognitive science. \textit{Behavioral and Brain Sciences}, 36(3), 181--204.

\bibitem{friston2010free}
Friston, K. (2010). The free-energy principle: A unified brain theory? \textit{Nature Reviews Neuroscience}, 11(2), 127--138.

\bibitem{zacks2007event}
Zacks, J. M., et al. (2007). Event perception: A mind-brain perspective. \textit{Psychological Bulletin}, 133(2), 273--293.

\bibitem{petroni2019language}
Petroni, F., et al. (2019). Language models as knowledge bases? In \textit{EMNLP}.

\bibitem{geva2021transformer}
Geva, M., Schuster, R., Berant, J., \& Levy, O. (2021). Transformer feed-forward layers are key-value memories. In \textit{EMNLP}.

\bibitem{power2022grokking}
Power, A., et al. (2022). Grokking: Generalization beyond overfitting on small algorithmic datasets. \textit{arXiv preprint arXiv:2201.02177}.

\bibitem{zhao2025probing}
Zhao, R., et al. (2025). Do we know what LLMs don't know? A study of consistency in knowledge probing. In \textit{Findings of EMNLP}.

\bibitem{miao2023selfcheck}
Miao, N., et al. (2023). SelfCheck: Using LLMs to zero-shot check their own step-by-step reasoning. In \textit{NeurIPS}.

\bibitem{wu2025rote}
Wu, Q., et al. (2025). Rote learning considered useful: Generalizing over memorized data in LLMs. In \textit{ICML 2025}.

\bibitem{tononi2006sleep}
Tononi, G., \& Cirelli, C. (2006). Sleep function and synaptic homeostasis. \textit{Sleep Medicine Reviews}, 10(1), 49--62.

\bibitem{fountas2024human}
Fountas, Z., et al. (2024). Human-like episodic memory for infinite context LLMs. \textit{arXiv preprint arXiv:2407.09450}.

\bibitem{lmlm2025}
Zhao, L., et al. (2025). Pre-training limited memory language models with internal and external knowledge. \textit{arXiv preprint arXiv:2505.15962}.

\bibitem{zelikman2022star}
Zelikman, E., Wu, Y., Mu, J., \& Goodman, N. (2022). STaR: Bootstrapping reasoning with reasoning. In \textit{NeurIPS}.

\bibitem{wang2023selfinstruct}
Wang, Y., et al. (2023). Self-Instruct: Aligning language models with self-generated instructions. In \textit{ACL}.

\bibitem{chen2024selfplay}
Chen, Z., et al. (2024). Self-play fine-tuning converts weak language models to strong language models. In \textit{ICML}.

\bibitem{bai2022constitutional}
Bai, Y., et al. (2022). Constitutional AI: Harmlessness from AI feedback. \textit{arXiv preprint arXiv:2212.08073}.

\bibitem{selftuning2024}
Zhang, X., et al. (2024). Self-Tuning: Instructing LLMs to effectively acquire new knowledge through self-teaching. \textit{arXiv preprint arXiv:2406.06326}.

\bibitem{wang2023selfconsistency}
Wang, X., et al. (2023). Self-consistency improves chain of thought reasoning in language models. In \textit{ICLR}.

\bibitem{atreja2025alas}
Atreja, D., et al. (2025). ALAS: Autonomous learning agent for self-updating language models. \textit{arXiv preprint arXiv:2508.15805}.

\bibitem{qu2025autonomous}
Qu, C., et al. (2025). Unlocking LLMs' self-improvement capacity with autonomous learning. In \textit{Findings of ACL}.

\bibitem{gao2025selfevolving}
Gao, J., et al. (2025). Self-evolving LLMs via continual instruction tuning. \textit{arXiv preprint arXiv:2509.18133}.

\bibitem{kirkpatrick2017overcoming}
Kirkpatrick, J., et al. (2017). Overcoming catastrophic forgetting in neural networks. \textit{PNAS}, 114(13), 3521--3526.

\bibitem{stable2025}
Bhoy, A., et al. (2025). STABLE: Gated continual learning for large language models. \textit{arXiv preprint arXiv:2510.16089}.

\bibitem{chekalina2024gate}
Chekalina, V., et al. (2024). SparseGrad: Efficient parameter-efficient fine-tuning via sparse gradient selection. \textit{arXiv preprint}.

\bibitem{nvidia2026ttt}
NVIDIA Research. (2026). Reimagining LLM memory: Using context as training data unlocks models that learn at test-time. \textit{NVIDIA Technical Blog}.

\bibitem{mcclelland1995complementary}
McClelland, J. L., McNaughton, B. L., \& O'Reilly, R. C. (1995). Why there are complementary learning systems in the hippocampus and neocortex. \textit{Psychological Review}, 102(3), 419--457.

\bibitem{yin2023selfknowledge}
Yin, Z., et al. (2023). Do large language models know what they don't know? In \textit{Findings of ACL}.

\bibitem{kadavath2022language}
Kadavath, S., et al. (2022). Language models (mostly) know what they know. \textit{arXiv preprint arXiv:2207.05221}.

\bibitem{kempner2024epistemic}
Ahdritz, G., et al. (2024). Distinguishing the knowable from the unknowable with language models. In \textit{ICML}.

\bibitem{hu2021lora}
Hu, E., et al. (2021). LoRA: Low-rank adaptation of large language models. In \textit{ICLR}.

\bibitem{touvron2023llama}
Touvron, H., et al. (2023). LLaMA: Open and efficient foundation language models. \textit{arXiv preprint arXiv:2302.13971}.

\bibitem{gekhman2024hallucinations}
Gekhman, Z., et al. (2024). Does fine-tuning LLMs on new knowledge encourage hallucinations? In \textit{EMNLP}.

\bibitem{ovadia2024finetuning}
Ovadia, O., et al. (2024). Fine-tuning or retrieval? Comparing knowledge injection in LLMs. In \textit{EACL}.

\end{thebibliography}
\end{document}